# Towards Physically Realizable Adversarial Attenuation Patch against SAR Object Detection

Yiming Zhang, Weibo Qin, Feng Wang

*Key Laboratory for Information Science of Electromagnetic Waves (MoE)*

*School of Information Science and Technology, Fudan University*

Shanghai 200433, China

zhangym21@m.fudan.edu.cn, wbqin23@m.fudan.edu.cn, fengwang@fudan.edu.cn

***Abstract*—Deep neural networks have demonstrated excellent performance in SAR object detection tasks, but remain susceptible to adversarial attacks. Existing SAR-specific attack methods can effectively deceive detectors while introducing noticeable perturbations to the digital domain, neglecting physical implementation constraints for attacking SAR object detection models. In this paper, a novel adversarial attack named Adversarial Attenuation Patch (AAP) is proposed, which employs an energy-constrained optimization strategy coupled with an attenuation-based deployment framework to achieve a seamless balance between attack effectiveness and stealthiness. AAP exhibits strong potential for physical realization by aligning with signal-level electronic jamming. Experimental results show that AAP effectively degrades detection performance while preserving high imperceptibility and shows favorable transferability across different models. This study provides a new perspective for physically realizable adversarial attacks against SAR object detection. The source code is made available at https://github.com/boremycin/SAAP.**

## I. Introduction

Synthetic aperture radar (SAR) is an active sensing system that leverages platform motion to synthesize a large virtual aperture and employs coherent signal processing to produce high-resolution images [1]. It possesses the unique capability to conduct observations under all-time and all-weather conditions [2]. With the rapid advancement of deep learning, deep neural networks (DNNs) have been extensively adopted for SAR image interpretation, especially in object detection tasks [3].

Despite their remarkable performance, DNNs are intrinsically vulnerable to adversarial examples, as demonstrated by Szegedy et al. [4], where carefully crafted perturbations can induce erroneous predictions. Motivated by this finding, numerous attack approaches have been proposed across diverse DNN-based tasks, demonstrating potent effectiveness. Conventional adversarial attacks distribute quasi-imperceptible perturbations across the entire image, whereas patch-based attacks localize perturbations and relax the imperceptibility constraint, improving practicality for real-world deployment [5]. In the SAR domain, recent work has sought to incorporate domain-specific characteristics into adversarial example generation [6, 7]. These approaches highlight the importance of integrating SAR-specific priors into adversarial attack design.

However, unlike Electro-Optical images, SAR images are typically grayscale and primarily characterized by sparse structures with limited semantic redundancy. As a result, directly applying image-domain patch-based attacks often introduces conspicuous artifacts that deviate from the underlying scattering mechanism, making the perturbations easily detectable and hindering the real-world deployment. This limitation stems from the fundamental mismatch between image-domain perturbations and the physical SAR imaging process. To tackle the issue, the SAR imaging mechanism is incorporated into a more physically viable adversarial patch design. Inspired by barrage jamming strategies [8, 9], we propose an Adversarial Attenuation Patch (AAP) that selectively suppresses target backscattered energy while constraining perturbation magnitude, improving both effectiveness and perceptual plausibility. In summary, the main contributions of this paper are as follows:

(1) A physically realizable patch-based attack method termed **Adversarial Attenuation Patch (AAP)** is proposed, where the adversarial patch is constrained in both location and magnitude to ensure compatibility with the feasible electronic jamming implementations.

(2) Extensive experiments on the SAR-Ship-Dataset [10] demonstrate that our method outperforms benchmark patch-based methods. Furthermore, the AAP maintains superior perceptual stealthiness, making adversarial examples significantly harder to identify as abnormal instances.

## II. Methodology

### A. Problem Formulation and Attack Objectives

For an input image $x \in \mathbb{R}^{H\times W}$, a DNN-based detector outputs bounding boxes $B_i$, class labels $C_i$ and confidence scores $S_i$, formulated as:

$$\hat{y} = f(x;\theta) = \{B_i, C_i, S_i\}_{i=1}^{N}, \tag{1}$$

where $f(\cdot;\theta)$ denotes a SAR detector with parameters $\theta$, and $\hat{y}$ is the prediction. The adversarial patch is obtained by solving the following optimization problem [11]:

$$arg\,\max_{\delta} \mathbb{E}_{(x,y)\sim D, t\sim T}\left[L\left(f\left(A(\delta,x,t)\right),y\right)\right], \tag{2}$$

where $\delta$ denotes the adversarial patch for input image $x$ with its annotations $y$ sampled from the dataset $D$. The term $t$ represents a stochastic transformation derived from the distribution $T$, which mainly encompasses random cropping, rotation, and brightness adjustment. $L(\cdot)$is the detection loss, and $A(\cdot)$ represents the patch application function.

### B. Attenuation-based Patch Deployment

The AAP method aims to generate a convergent adversarial patch through an iterative gradient-based optimization process, which can be formulated as:

$$\delta_{i+1} = \delta_i + \alpha \cdot sign\left(\nabla_\delta L\left(f\left(A\left(\delta_i, x_i^{adv}, t\right)\right), y\right)\right), \tag{3}$$

where $\delta_i$ and $x_i^{adv}$ represent the patch and the corresponding adversarial example at the i-th respectively. The term $\nabla_\delta L(\cdot)$ is the gradient of the loss with respect to the patch, which is non-zero only within the patch-located region, while $\alpha$ is the step size.

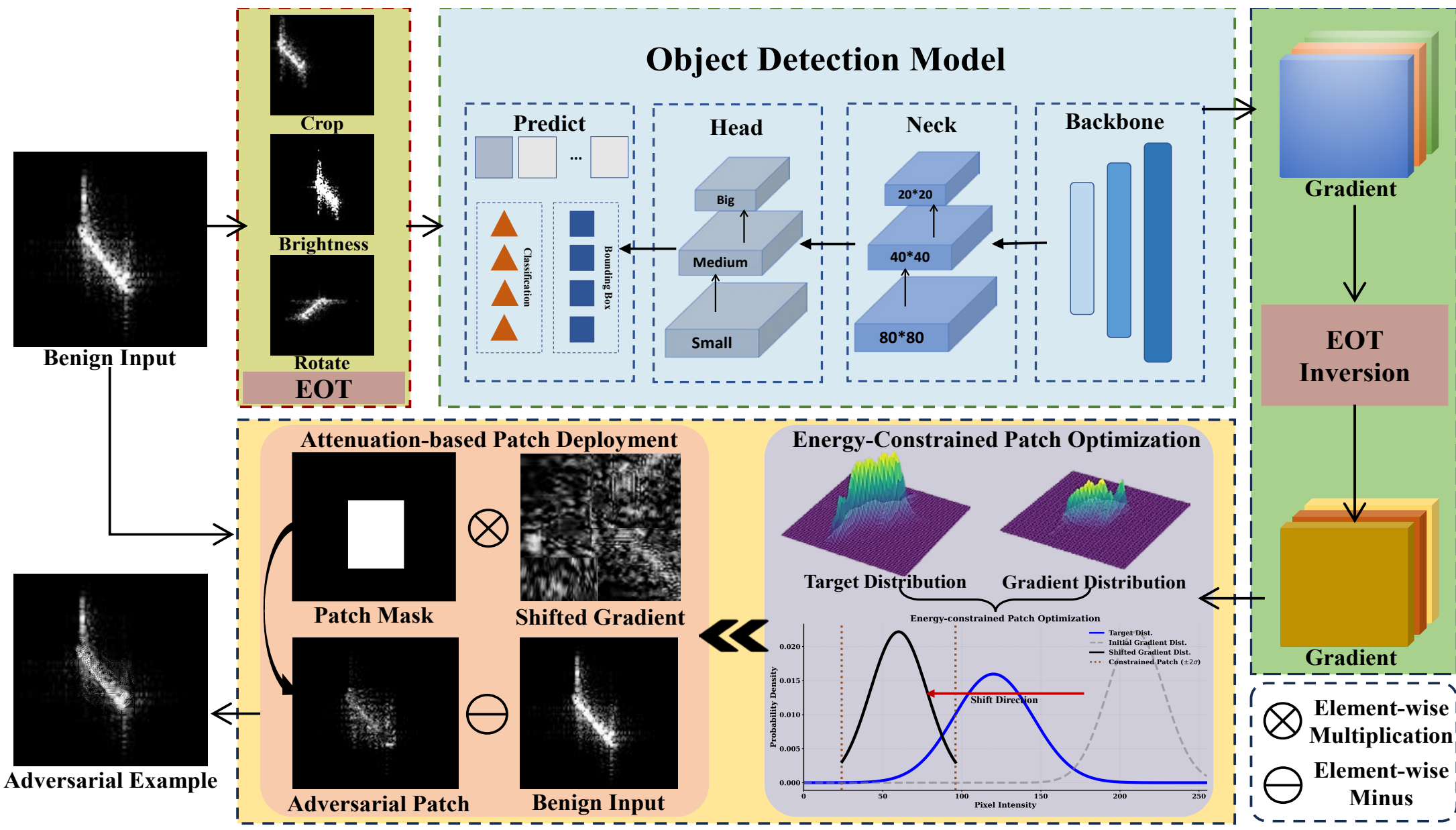

**Fig. 1. Pipeline of the proposed physically realizable attenuation patch attack.**

In the deployment phase, the conventional procedure directly replaces the patch region utilizing a binary mask $M$. Given a patch $\delta \in \mathbb{R}^{C\times h\times w}$ located at $(x_p, y_p)$, the adversarial example is generated as:

$$x^{adv} = x \odot (1 - M) + \delta \odot M, \tag{4}$$

where $\odot$ represents the Hadamard product. The binary matrix $M$ is constructed according to the patch region $\Omega = [x_p, x_p + h] \times [y_p, y_p + w]$ and is formulated as:

$$M(i,j) = \begin{cases} 1, & (i,j) \in \Omega \\ 0, & otherwise. \end{cases} \tag{5}$$

The notable discrepancy in backscattering properties between targets and ambient clutter typically results in higher pixel intensities within target areas, making direct patch replacement easily detectable. Therefore, an attenuation-based deployment strategy is proposed to exploit target saliency and high-density regions. The adversarial patch is applied via subtractive modulation, preserving structural and intensity consistency while better approximating realistic SAR jamming scenarios. This process is formulated as:

$$x^{adv} = x - \delta \odot M. \tag{6}$$

From a physical perspective, the implementation of AAP can be interpreted as a controlled form of SAR active concealment jamming. Unlike conventional strategies that aim to completely mask the receiving signals within the Region of Interest (ROI), the realization of AAP is envisioned to modulate the jamming signal to achieve a specific attenuation effect. The resulting image, reconstructed from the superimposed echoes, consistently matches the corresponding digital adversarial patch within the ROI. The whole process can be formulated as:

$$x^{phy} = I\left(s_r(t_r, t_a) + s_j(t_r, t_a)\right), \tag{7}$$

where $x^{phy}$ represents the physically reconstructed adversarial example. $I(\cdot)$ signifies the SAR imaging algorithm, $s_r$ and $s_j$ represent the received echoes from the scene and the injected jamming signals, respectively, while $t_r$ and $t_a$ denote the fast-time and slow-time during the imaging process.

## C. *Energy-Constrained Patch Optimization*

Based on the above discussion, the perturbation magnitude of the patch requires an elaborate design to guarantee imperceptibility while maintaining attack effectiveness. Since the attenuation-based deployment diminishes the target area, the original pixel intensity and distribution should be meticulously considered.

To address this issue and align patch intensity with the target region, an adaptive energy constraint is introduced to regulate the adversarial patch during optimization. Specifically, we compute the mean intensity of the original image over the patch region $\Omega$, which is defined as:

$$E_x = \frac{1}{h\cdot w} \sum_{(i,j)\in\Omega} x_{ij}\,. \tag{8}$$

Furthermore, the target energy level of the adversarial patch is defined as follows:

$$E_p = \tau \cdot E_x, \tag{9}$$

where $\tau$ denotes a scaling coefficient that modulates the relative intensity contrast between the patch and the benign target region, thereby ensuring the energetic alignment.

To accommodate diverse intensity distributions across miscellaneous target instances, the patch adjusts its mean intensity while preserving spatial structure to maintain the adversarial effectiveness, which ultimately dictates the following optimization strategy:

$$\delta' = \delta - \mu_\delta + E_p. \tag{10}$$

For further stabilizing adversarial patterns and suppressing artifacts or outliers, a dynamic clipping function is applied to constrain the intensity fluctuations:

$$\delta^{adv} = clip(\delta', E_p - 2\sigma_\delta, E_p + 2\sigma_\delta), \tag{11}$$

where $\mu_\delta$ and $\sigma_\delta$ represent the mean and standard deviation of the patch, respectively.

Table 1. Adversarial effectiveness comparison of patch-based attacks on YOLOv5-M as the white-box model. The best results are **bold** and the second best are underlined.

| Attack | mAP50↓ | mAP50-95↓ | FNR↑ | FPPI↑ |
|---|---|---|---|---|
| Clean | 97.9 | 69.1 | 0.040 | 0.043 |
| DPatch | 90.4 | 65.8 | 0.065 | 0.052 |
| RobustDPatch | 92.2 | 67.5 | 0.087 | 0.143 |
| AdvPatch | 64.0 | 48.8 | 0.304 | 0.668 |
| **AAP (Ours)** | **37.2** | **16.2** | **0.413** | **1.021** |

### D. *Final Adversarial Example Generation*

During generation, the application function $A(\cdot)$ consists of two stages: transformation-based augmentation and attenuation-based deployment.

For transformation augmentation, the Expectation over Transformation (EOT) [12] strategy is adopted. At each iteration, a set of random transformations $t(\cdot)$, which are sampled from a predefined distribution $T$ and include random cropping, rotation and brightness adjustment, are applied to $x$. This process enables the computation of gradients that are more robust and consistent with real-world scenarios.

The patch is then iteratively updated according to Eq. (3), while the energy constraint in Eqs. (10)-(11) is incorporated to regulate its intensity distribution. The function $A(\cdot)$ unifies both transformation-based augmentation and energy-constrained patch optimization, and is formulated as:

$$A(\delta, x, t) = t^{-1}(t(x) - \delta \odot M), \tag{12}$$

where $t^{-1}(\cdot)$ denotes the inverse transformation and $M$ is the binary mask indicating the patch region.

After obtaining the optimized adversarial patch $\delta^{adv}$, the final adversarial example is generated via the attenuation-based deployment defined in Eq. (6). This formulation provides a more realistic modeling of physical signal interference and significantly improves the stealthiness of the attack. The overall workflow of our proposed AAP is depicted in Fig. 1.

## III. EXPERIMENTS

### A. *Experimental Setup*

The experiments are conducted on the SAR-Ship-Dataset [10], which contains 39729 ship instances with a resolution of $256 \times 256$. We adopt YOLOv5 [13] and YOLOv11 [14] as detection models, with YOLOv5-M serving as the white-box model. To evaluate transferability, multiple unseen variants, including YOLOv5-L/X and YOLOv11-N/M/X, are used for black-box testing. We compare AAP with three patch-based attacks, namely AdvPatch [15], DPatch [16], and RobustDPatch [11]. AdvPatch is adapted from classification tasks, while DPatch and RobustDPatch are inherently tailored for Electro-Optical object detection.

For fair comparison, 100 test images are randomly selected, and all methods are optimized for 3,000 iterations with a learning rate of $\alpha = 5.0$. AdvPatch uses a patch covering 15% of the input image, while DPatch and RobustDPatch adopt a fixed size of $120 \times 120$. All baseline patches are randomly placed, whereas AAP determines patch size and location based on the ground-truth labels. For EOT, the crop size, brightness range, and rotation probabilities are set to [50, 50], [0.7, 1.4], and [0.4, 0.2, 0.2, 0.2], respectively.

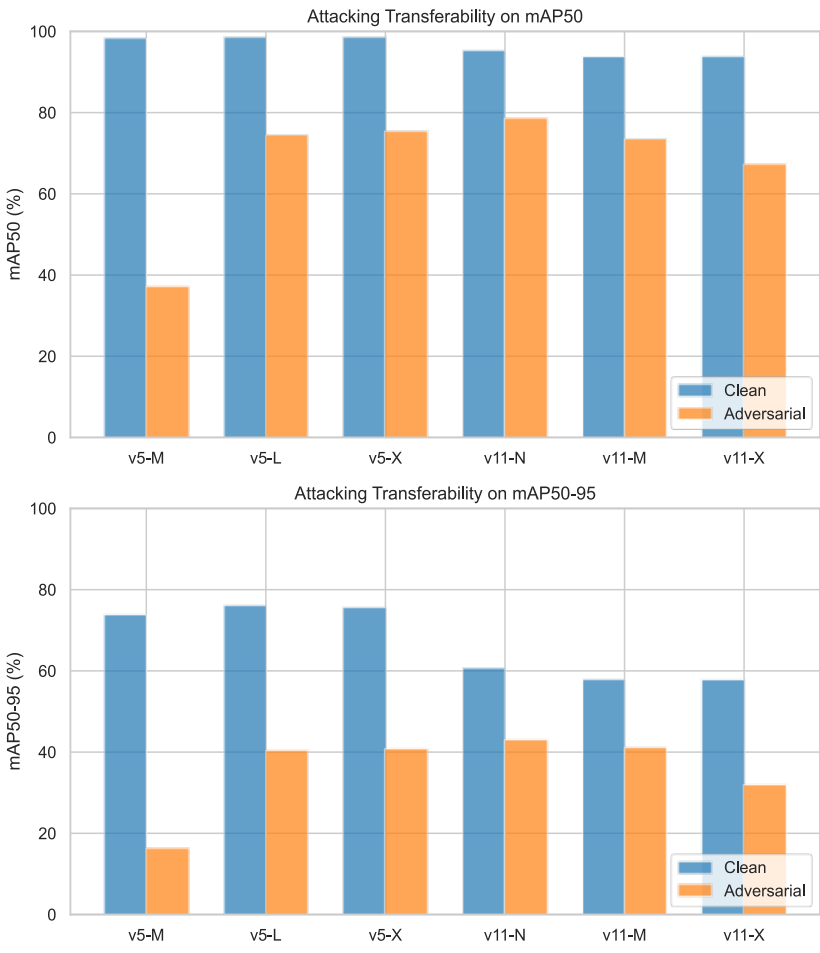

**Fig. 2. Transferability of adversarial examples generated by YOLOv5-M across different models, measured by mAP50 and mAP50–95 under clean and adversarial settings.**

### B. *Evaluation Metrics*

To quantitatively assess the detection performance under the attack of adversarial examples, mean Average Precision (mAP) is primarily utilized. Decreases in mAP, including mAP50 and mAP50–95, reflect performance degradation induced by adversarial examples, capturing heterogeneous failure modes such as target vanishing, false alarm, and location offset at the whole dataset scale.

In addition, False Negative Rate (FNR) and False Positives Per Image (FPPI) are adopted as complementary metrics. FNR measures the proportion of missed detections over all ground-truth objects, while FPPI quantifies the density of false alarms per image, reflecting instance-level detection noise.

### C. *Adversarial Effectiveness and Transferability*

Table 1 reports the performance of AAP against existing patch-based attacks. The YOLOv5-M detector serves as the white-box model, achieving 96.7% mAP50 and 69.1% mAP50-95, providing a strong baseline for attack evaluation.

AAP consistently outperforms all competing methods. It reduces mAP50 and mAP50-95 by **26.8%** and **32.6%** more than the second-best AdvPatch, respectively, while achieving the highest FNR (0.413) and FPPI (1.021), surpassing AdvPatch by margins of **0.109** and **0.352**.

As shown in Fig. 2, adversarial examples generated from the surrogate YOLOv5-M transfer effectively to unseen models, including YOLOv5-L/X and YOLOv11-N/M/X, demonstrating strong cross-architecture transferability.

These results indicate that the proposed optimization and deployment strategies enable the learned perturbation to capture model-agnostic vulnerabilities, highlighting the practical potential in physical deployment scenarios.

### D. *Visualization and Effect Analysis*

A preliminary taxonomy of adversarial effects induced by AAP is presented. As shown in Fig. 3, the attacks can be categorized into three failure modes: **confidence drop**, where object instances are detected with pronounced low confidence; **false alarm**, where numerous spurious detections are induced, seriously disrupting the normal interpretation; and **missed detection**, where target objects are completely vanished from the detector. Fig. 5 further shows the distribution of these failure modes over the set. The results demonstrate that our proposed AAP can cause diverse failure modes in object detection, enhancing its effectiveness for realistic applications.

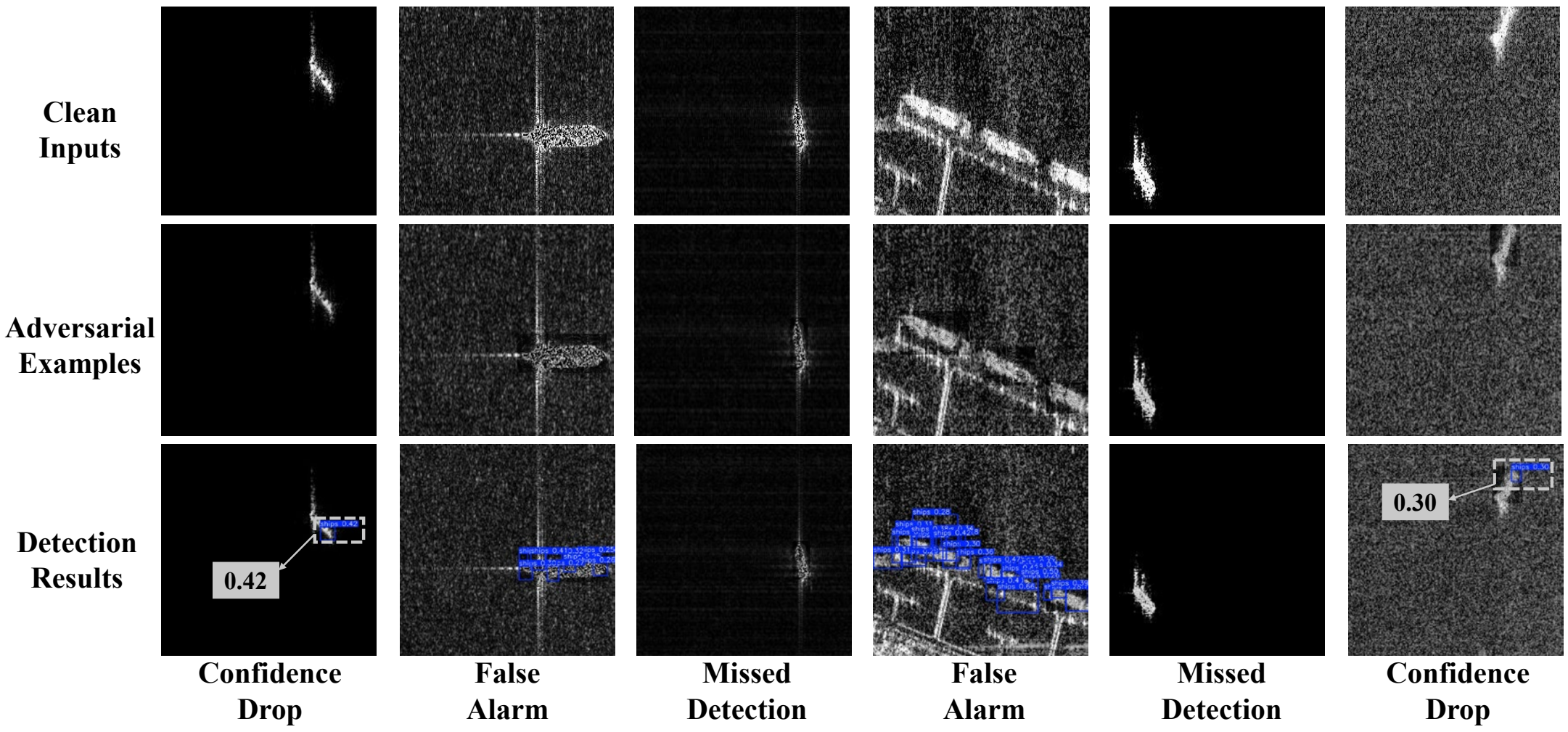

**Fig. 3. Visual comparison of clean inputs, generated adversarial examples, and their corresponding detection results. Three representative attack effects are illustrated: confidence drop, false alarm, and missed detection, demonstrating the degradation of detection reliability.**

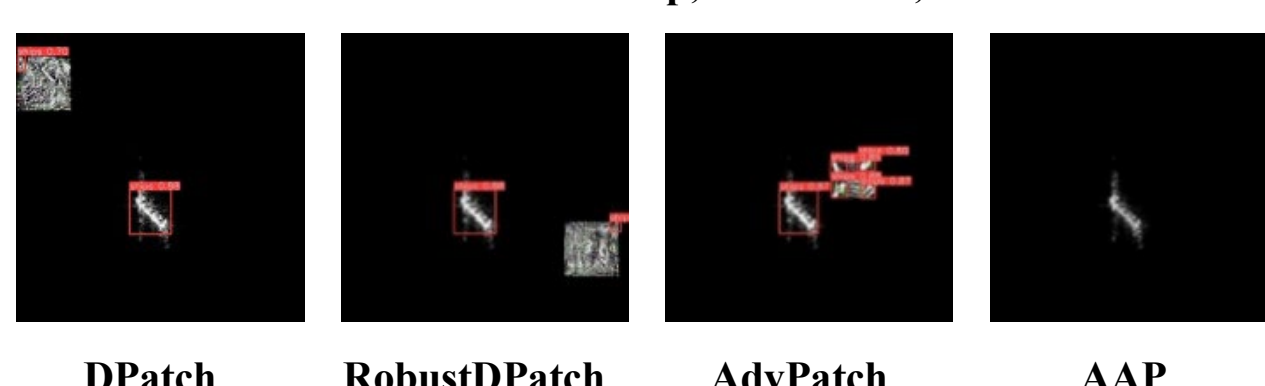

**Fig. 4. Visualization of adversarial examples generated by different methods, with corresponding detection bounding boxes.**

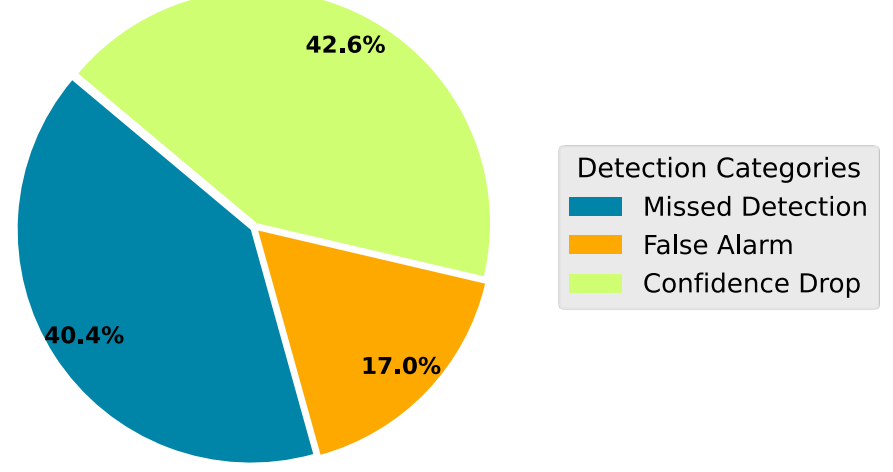

**Fig. 5. Proportional distribution of failure modes induced by AAP.**

Fig. 4 represents a visual comparison between adversarial examples generated by different methods and their corresponding detection results. Owing to the SAR-specific optimization and deployment strategies, AAP achieves strong perceptual stealthiness without introducing obvious artifacts in background regions, while still significantly degrading detection reliability.

## IV. Conclusion

In this work, a novel adversarial patch termed AAP is proposed, which is tailored for SAR object detection with potential for physical realization. By integrating an energy-constrained optimization strategy with an attenuation-based deployment framework, our approach achieves significant drops in mAP while maintaining high perceptual imperceptibility compared with patch replacement methods. Experimental results reveal that AAP effectively induces diverse failure modes, including missed detection, false alarm, and confidence drop. Specifically, AAP demonstrates robust cross-scale and cross-architecture adversarial transferability, facilitating its operational utility for real-world applications.